\pdfoutput=1

\documentclass[11pt]{article}
\usepackage{booktabs}
\usepackage{amsmath}
\usepackage[breaklinks=true]{hyperref}
\usepackage[preprint]{acl}

\usepackage{times}
\usepackage{latexsym}
\usepackage{enumitem}
\usepackage{placeins}
\usepackage{tcolorbox}

\usepackage{xurl}

\usepackage[T1]{fontenc}

\usepackage[utf8]{inputenc}

\usepackage{microtype}

\usepackage{inconsolata}

\usepackage{graphicx}

\title{GeoChain: Multimodal Chain-of-Thought for Geographic Reasoning}

\author{Sahiti Yerramilli\thanks{Equal Contribution.} \\
  Google \\
  \texttt{sahitiy@google.com}\\\And
  Nilay Pande\footnotemark[1] \\
  Waymo \\
  \texttt{nilayp@waymo.com} \\\AND
  Rynaa Grover\footnotemark[1] \\
  Google \\
  \texttt{rynaa@google.com} \\\And
  Jayant Sravan Tamarapalli\footnotemark[1] \\
  Google \\
  \texttt{jayantsravan@google.com}}

\makeatletter
\renewcommand\@fnsymbol[1]{\ensuremath{\ifcase#1\or \dagger\or \ddagger\or \mathsection\or \mathparagraph\or \|\or **\or \dagger\dagger\or \ddagger\ddagger\else\@ctrerr\fi}}

\begin{document}
\maketitle
\begin{abstract}
This paper introduces GeoChain, a large-scale benchmark for evaluating step-by-step geographic reasoning in multimodal large language models (MLLMs). Leveraging 1.46 million Mapillary street-level images, GeoChain pairs each image with a 21-step chain-of-thought (CoT) question sequence (over 30 million Q\&A pairs). These sequences guide models from coarse attributes to fine-grained localization across four reasoning categories - visual, spatial, cultural, and precise geolocation - annotated by difficulty. Images are also enriched with semantic segmentation (150 classes) and a visual locatability score. Our benchmarking of contemporary MLLMs (GPT-4.1 variants, Claude 3.7, Gemini 2.5 variants) on a diverse 2,088-image subset reveals consistent challenges: models frequently exhibit weaknesses in visual grounding, display erratic reasoning, and struggle to achieve accurate localization, especially as the reasoning complexity escalates. GeoChain offers a robust diagnostic methodology, critical for fostering significant advancements in complex geographic reasoning within MLLMs. 

\textbf{Code:}~\href{https://github.com/sahitiy/geochain}{\url{https://github.com/sahitiy/geochain}}

\textbf{Dataset:}~\href{https://huggingface.co/datasets/sahitiy51/geochain}{\url{https://huggingface.co/datasets/sahitiy51/geochain}}
\end{abstract}

\section{Introduction}
\begin{figure*}[t!] 
    \centering
    \includegraphics[width=0.7\textwidth]{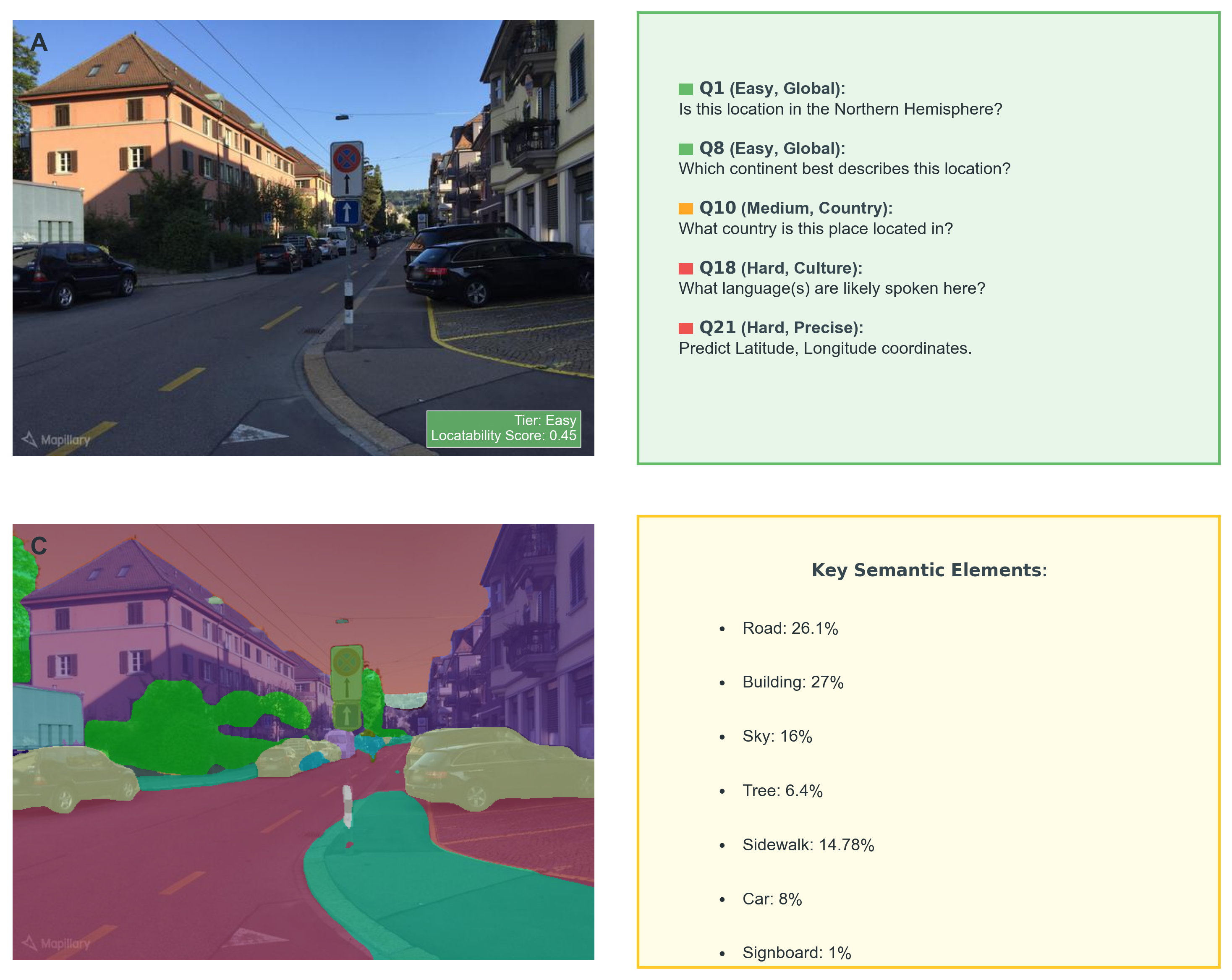}
    \caption{Components of a GeoChain instance: \textbf{(Top-Left)} Easy Mapillary Street-Level Sequences (MSLS) image with locatability score of 0.45. \textbf{(Top-Right)} Example chain-of-thought questions with difficulty indicators. \textbf{(Bottom-Left)} Derived semantic segmentation map. \textbf{(Bottom-Right)} Extracted key semantic labels. Together, these elements enable step-by-step diagnostic evaluation of geographic reasoning.}
    \label{fig:overview}
\end{figure*}

As large vision-language models (VLMs) continue to make rapid progress on general visual question answering and captioning tasks \cite{geminiteam2024gemini15unlockingmultimodal, openai2024gpt4technicalreport, wang2024qwen2vlenhancingvisionlanguagemodels, dai2023instructblipgeneralpurposevisionlanguagemodels}, their capacity for structured geographic reasoning remains underexplored. The ability to infer a location from visual cues -such as terrain, signage, vehicles, or architecture - considered alongside spatial and cultural knowledge, is crucial for real-world applications like remote sensing, disaster response, and autonomous navigation. More broadly, geographic localization serves as a testbed for grounded intelligence, requiring models to reason over subtle visual features, incorporate world knowledge, and disambiguate locations that may be visually similar. Despite this, existing benchmarks rarely probe the kind of step-by-step reasoning that such tasks demand.

We introduce GeoChain, a novel multimodal benchmark for evaluating structured geographic reasoning in large language models (MLLMs). As depicted in Figure \ref{fig:overview}, each GeoChain sample features a street-level image from the Mapillary dataset \cite{Warburg2020MapillarySL} paired with a 21-step chain-of-thought (CoT) question sequence. These sequences progressively guide models from coarse inferences, such as hemisphere or continent, to fine-grained predictions like city, latitude, and longitude. The complete GeoChain framework comprises 1.46 million images, each with this 21-question CoT structure, yielding over 30 million question-answer pairs. Questions span four core reasoning categories: visual cues, spatial localization, cultural inference, and precise geolocation, all annotated with difficulty levels for granular evaluation. This curriculum-style structure offers vital diagnostic insights into where and why models fail across reasoning stages, moving beyond sole reliance on final predictions.

To facilitate focused evaluations, we curated GeoChain Test-Mini, a diverse and challenging subset. This curation process leverages a locatability score, adapted from GeoReasoner \cite{li2024georeasoner} and computed using features from a pretrained MaskFormer model \cite{Cheng2021MaskFormer}. This score quantifies the visual identifiability of a location from a single image, allowing us to stratify GeoChain Test-Mini into Easy, Medium, and Hard tiers based on thresholds in the 0.12-0.6 range. The resulting GeoChain Test-Mini contains 2,088 carefully selected images, designed to offer a representative yet manageable scale for robust MLLM assessment.

Our main contributions are as follows:
\begin{itemize} [itemsep=5pt, leftmargin=5pt, labelsep=7pt, parsep=2pt]
    \item \textbf{GeoChain Benchmark Framework:} A novel benchmark for evaluating step-by-step MLLM geographic reasoning, derived from 1.46 million Mapillary street-level images and generating over 30 million Q\&A pairs through 21-step chain-of-thought questions, all structured across diverse reasoning categories and difficulty levels.
    \item \textbf{Rich Augmentation \& Curated Evaluation Set:} A methodology for enhancing images with semantic labels (150 classes) and a human-inspired locatability score for difficulty stratification, culminating in the GeoChain Test-Mini: a quality-controlled 2088-image evaluation set; the resulting rich semantic metadata also offers a valuable resource for broader community research and future investigations.
    \item \textbf{Comprehensive MLLM Benchmarking \& Analysis:} Evaluation of leading MLLMs on GeoChain Test-Mini, providing detailed insights into their geographic reasoning capabilities, performance variations, and common failure modes. 
\end{itemize}
\section{Related Work}

\subsection{Image-Based Geolocation}
Early work in visual geolocation predominantly focused on matching query images to large, geotagged image databases, often aiming for direct coordinate prediction. For instance, Im2GPS \cite{Hays2008Im2GPS} pioneered retrieving locations by comparing against a massive photograph dataset. Later, deep learning significantly advanced the field; PlaNet \cite{Weyand2016PlaNet} utilized convolutional neural networks (CNNs) for global location prediction, and architectures like NetVLAD \cite{arandjelovic2016netvlad} learned robust image representations for effective place recognition, improving upon earlier retrieval methods. Other approaches, such as those focusing on urban or cross-view settings \cite{8099699}, further specialized these techniques. GeoChain diverges from these paradigms, which primarily target endpoint localization accuracy or image retrieval. Instead, it introduces a structured multimodal reasoning benchmark where models must articulate a 21-step chain-of-thought (CoT) sequence of answers to geographically relevant questions, thereby enabling finer-grained diagnostic insight into their internal reasoning processes.

\subsection{Multimodal Geographic Reasoning and Benchmarks}
More recent efforts have begun to integrate visual understanding with language-based reasoning for complex geographic tasks. GeoReasoner \cite{li2024georeasoner}, for example, introduced a fine-tuning strategy for MLLMs using human gameplay traces, primarily to improve final location prediction by modeling human-like inference. Similarly, other recent studies \cite{pramanik2024evaluatingprecise, yang2024vlmsgeoguessr} also concentrate on predicting precise latitude and longitude. GeoComp \cite{song2025geolocationrealhumangameplay} presents a large-scale dataset of geolocation gameplay data, emphasizing step-wise reasoning rooted in real human gameplay that often involves external metadata, active exploration, and dynamic information gathering. While these approaches offer valuable insights into human-like inference and gameplay dynamics, GeoChain's contribution is complementary. It does not involve model fine-tuning or rely on gameplay trajectories. Instead, GeoChain employs a fully static, image-grounded evaluation framework: each sample consists of a single image paired with its fixed CoT question sequence, standardized across the entire benchmark. This design facilitates direct and controlled benchmarking of different models' inherent reasoning capabilities under uniform conditions, distinct from evaluating exploratory strategies or the ability to process dynamic data.

Other benchmarks, such as GAEA \cite{campos2025gaea}, generate diverse conversational questions from detailed, place-specific metadata like OpenStreetMap attributes. While this can create rich contextual queries, it introduces challenges related to the temporal stability of dynamic data (e.g., changes in urban landscape) and complicates fair, apples-to-apples model comparisons due to non-uniform question sets. Consequently, disentangling model reasoning failures from idiosyncratic question characteristics becomes difficult. GeoChain mitigates these issues by grounding its standardized questions in more enduring visual semantics such as the presence of characteristic vegetation, architectural styles, or road infrastructure, often identifiable through image segmentation and stable general geographic facts. This focus ensures the evaluation centers on the consistency of the reasoning process itself.

Furthermore, existing geospatial benchmarks like GEO-Bench \cite{lacoste2023geobenchfoundationmodelsearth} primarily target remote sensing applications, offering valuable tools for Earth monitoring with satellite imagery and tasks such as classification or segmentation. In contrast, GeoChain specifically addresses agent-level geographic reasoning from high-resolution, ground-level imagery, emphasizing natural-language understanding of spatial, cultural, and visual cues directly perceivable in such environments.

\subsection{Mapillary Street-Level Sequences Dataset}
GeoChain is built upon the Mapillary Street-Level Sequences (MSLS) dataset \cite{Warburg2020MapillarySL}, a large-scale, crowd-sourced collection of diverse, geo-tagged street-level images. MSLS's global coverage, with data from numerous cities worldwide reflecting the breadth of the MSLS ecosystem, and its varied capture conditions (diverse cameras, viewpoints, seasons, times of day) make it an ideal foundation for a benchmark aiming to evaluate generalizable geographic reasoning.

\section{GeoChain Benchmark Construction}
\label{sec:method}

\begin{figure*}[t!]
 \centering
 \includegraphics[width=0.9\textwidth]{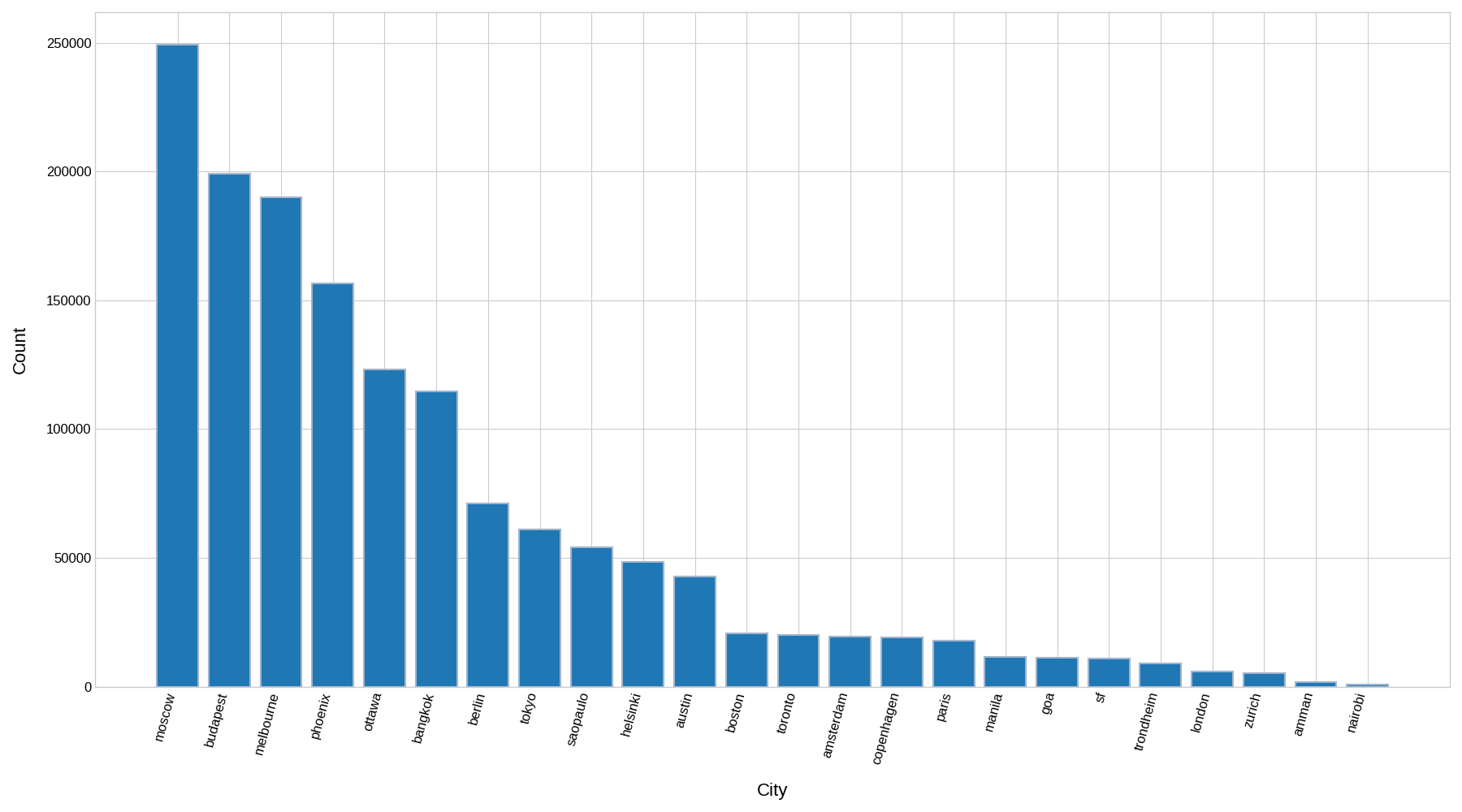}
 \caption{Count of images per city, illustrating the city distribution within the GeoChain dataset.}
 \label{fig:data_distr}
\end{figure*}

The GeoChain benchmark is constructed by augmenting the Mapillary Street-Level Sequences (MSLS) dataset \cite{Warburg2020MapillarySL}. MSLS provides a diverse collection of geo-tagged street-level imagery (approximately 1.4 million images in its full extent, with a geographical distribution across numerous cities as illustrated in Figure \ref{fig:data_distr}), crucial for developing and evaluating geographic localization models. However, to facilitate fine-grained, step-by-step reasoning, we introduce several layers of annotation and metadata. Our contributions enhance the MSLS dataset in three primary ways: semantic class labeling, locatability score computation, and the design of a structured chain-of-thought question battery. These augmentations, followed by a careful test set curation process, collectively enable a more nuanced evaluation of multimodal models' geographic reasoning capabilities.

\subsection{Semantic Class Labeling}
\label{ssec:class_labeling}
To ground visual reasoning in explicit semantic content, each image in our benchmark is augmented with a semantic segmentation map. This map provides a detailed understanding of the scene's composition by identifying various objects and environmental features. We employed MaskFormer \cite{Cheng2021MaskFormer}, a state-of-the-art transformer-based architecture for semantic segmentation. Specifically, we utilized a MaskFormer model pre-trained on the ADE20K dataset \cite{Zhou2017ADE20KSceneParsing}, which offers a rich label set of 150 distinct classes, encompassing a wide array of objects, environmental elements (e.g., ``tree'', ``sky'', ``road''), and architectural features (e.g., ``building'', ``window'', ``door'').

From the segmentation map, we calculate how much of the image is covered by each category. We do this by working out the percentage of the image's total area that each specific category takes up. For example, we might find that `sky' covers 30\% of an image, and `road' covers 15\%. This measurement of what's in the scene, and how much of it there is, then helps us create the correct answers for many of the visual questions in our benchmark.

\subsection{Locatability Score Computation}
\label{ssec:locatability_score}
\begin{figure}[h]
 \centering
 \includegraphics[width=1\linewidth]{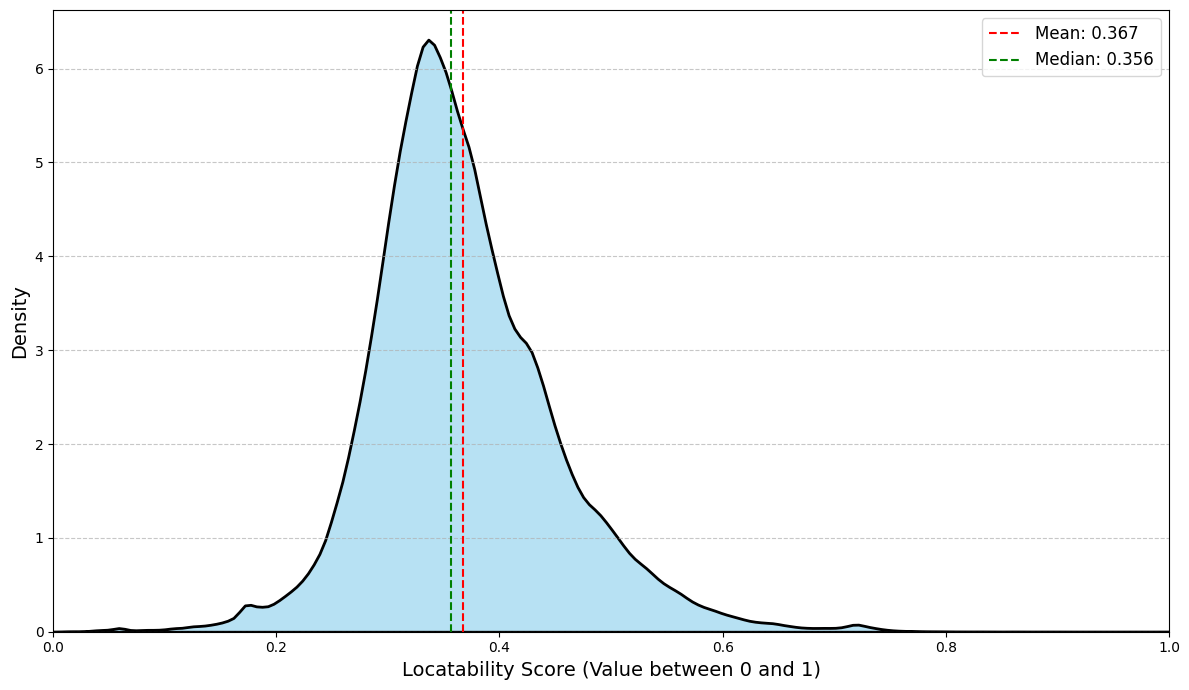}
\caption{\centering Distribution of Locatability Scores in GeoChain}
 \label{fig:loc_density}
\end{figure}

To systematically assess model performance across varying levels of visual ambiguity, we compute a \textit{locatability score} for every image considered for the GeoChain benchmark. This score, ranging from 0 to 1, quantifies how visually identifiable a location is likely to be, with higher scores indicating more distinct and easily locatable scenes. Our methodology for calculating this score is adopted from \cite{li2024georeasoner}. The distribution of these computed locatability scores across the considered images is shown in Figure \ref{fig:loc_density}.

The core idea behind this score is to leverage common visual cues that humans, particularly proficient GeoGuessr players \cite{geoguessr}, rely on for geolocalization. The process involves several steps:
\begin{enumerate}[itemsep=1pt, leftmargin=10pt]
\item \textbf{Identification of Cues:} A set of cues frequently used by GeoGuessr players (e.g., “houses in central Chile are more likely to
have terracotta tiled roofs”) is established.
\item \textbf{Cue-to-Class Similarity:} The semantic similarity between these cues and the 150 class labels produced by the MaskFormer model (as described in Section \ref{ssec:class_labeling}) is computed. This typically involves using text embeddings to represent both the cues and the class labels, followed by a similarity measure (e.g., cosine similarity).
\item \textbf{Class Weight Derivation:} The similarities are aggregated across all cues for each class and then subjected to min-max normalization to derive a set of weights $w_c$ for each class c. These weights reflect the importance of each visual class for geolocalization.
\item \textbf{Weighted Score Aggregation:} The final locatability score for an image is computed as a weighted sum of the percentage areas of the classes present in the image.
\end{enumerate}

Using this score, we divide GeoChain images into three difficulty tiers:  \textbf{Hard} if $\text{score} \in [0.12, 0.22)$, \textbf{Medium} if $\text{score} \in [0.22, 0.45)$, and \textbf{Easy} if $\text{score} \in [0.45, 0.6)$. This stratification, based purely on visual cues inherent in the imagery, allows for a more granular analysis of model performance and helps identify specific weaknesses in reasoning about visually challenging environments.

\subsection{Chain-of-Thought Question Design}
\label{ssec:cot_questions}
A central component of GeoChain is a carefully designed sequence of 21 questions that guide the model through a step-by-step reasoning process, from coarse-grained observations to fine-grained localization. This chain-of-thought (CoT) approach aims to mimic a structured human-like deduction process. The questions are ordered such that earlier questions elicit information or focus attention on attributes that can be instrumental in answering subsequent, more complex questions.

The full list of 21 questions, along with their rank, assigned difficulty (Easy, Medium, Hard), question type (Binary, Multiclass, Free-text), and question category (e.g., Culture/Infrastructure, Geo Localization, Terrain/Environment), is provided in Appendix \ref{sec:questions}. The difficulty annotation (Easy, Medium, Hard) for each question reflects the anticipated challenge of answering that specific question in isolation, based on the type of information required.

The question set is designed to be static across all data points in the benchmark. This uniformity ensures a consistent evaluation framework, allowing for direct, apples-to-apples comparisons of different models' reasoning capabilities. The questions cover diverse aspects:
\begin{itemize} [itemsep=1pt, leftmargin=10pt]
\item \textbf{Visual Object/Attribute Presence:} Some questions directly query the presence of specific objects or attributes identifiable from the image (e.g., "Do you see any boats or ships?''). Ground truth for these questions is primarily derived from the semantic class labels extracted via the MaskFormer model (Section \ref{ssec:class_labeling}). For instance, if the class "boat'' occupies a non-zero percentage of the image, the answer would be affirmative.
\item \textbf{Inferential and Contextual Knowledge:} Other questions require more derivative reasoning or contextual knowledge beyond direct object identification (e.g., "Is this place near a coast?'', "What side of the road do vehicles drive on here?''). The MSLS dataset encompasses images from 24 distinct cities globally. For images originating from these locations, we manually curated ground truth answers for city-level attributes or environmental characteristics (e.g., predominant architectural styles, typical climate indicators) that apply broadly to the image's geographic area.
 \item \textbf{Progressive Localization:} The sequence progresses from general observations (e.g., hemisphere, continent) to specific details (e.g., country, city, precise latitude and longitude coordinates).
\end{itemize}
The question types include binary (Yes/No), multiclass (selection from a predefined set of options), and free-text (open-ended answers, such as country name or coordinates). This variety tests different aspects of a model's understanding and generation capabilities.

The semantic segmentation labels generated in Section \ref{ssec:class_labeling} were instrumental in constructing several questions that directly probe the visual understanding capabilities of the models. Beyond their use in the current benchmark, this rich semantic metadata, now part of GeoChain, offers a valuable resource for the community. It can be leveraged to design new questions aimed at further investigating specific aspects of model behavior, such as tendencies towards visual hallucination \cite{li2023evaluating} \cite{rohrbach-etal-2018-object} or the fine-grained ability to identify a wider array of objects. The insights derived from such extended evaluations can subsequently guide targeted improvements in model development.

By analyzing model performance across this structured chain of questions, GeoChain aims to provide deeper insights into the strengths and weaknesses of multimodal geographic reasoning systems.

\subsection{Test Set Curation and Sampling Strategy}
\label{ssec:test_set_curation}
To create the "GeoChain Test-Mini" subset for focused evaluation, we prioritized stratification, visual quality, and diversity. We initially targeted 2100 images, stratified by locatability scores into 700 Easy, 700 Medium, and 700 Hard examples. A tiered, unique-sequence sampling strategy was employed: unique image sequences were randomly sampled first for the Hard tier, then for the Medium tier (from remaining unique sequences), and finally for the Easy tier, ensuring no sequence was reused across tiers. The underlying MSLS dataset exhibits a notable skew in its per-city image distribution (as highlighted by the overall dataset statistics in Figure \ref{fig:data_distr}). Consequently, to avoid introducing new biases that could arise from attempting to manually balance city representation or 'carefully' over/under-sample from specific locations, our approach was to randomly sample unique image sequences across all available cities within each defined locatability tier. These 2100 candidates underwent manual visual inspection, where 12 images with critical quality issues (e.g., excessive blur, poor exposure) were removed. This rigorous curation yielded a final Test-Mini set of 2088 high-quality, diverse, and appropriately challenging images. 

\section{Analysis}
In this section, we evaluate the performance of frontier vision-language models: GPT-4.1, GPT-4.1-mini \cite{openai2024gpt4technicalreport}, Claude 3.7 Sonnet \cite{claude37sonnet}, Gemini 2.5 Flash \cite{google_gemini_2.5_flash_blog} and Gemini 2.5 Pro \cite{gemini2.5pro} on the GeoChain "Test-Mini" benchmark, focusing on their ability to reason accurately and consistently across a structured 21-step geographic reasoning chain.

\begin{table*}[htpb]
\centering
\setlength{\tabcolsep}{3pt}
\caption{Overall model-level accuracy and localization metrics.}
\label{tab:model_level}
\begin{tabular}{lccccc}
\toprule
\textbf{Model} & \textbf{Pass Score (\%)} & \textbf{Mean Dist (km)} & \textbf{< 25 km (\%)} & \textbf{< 200 km (\%)} & \textbf{< 750 km (\%)} \\
\midrule
Gemini-2.5-pro      & \textbf{81.84} & 489.51  & \textbf{59.38} & 69.95 & 88.51\\
Gemini-2.5-Flash    & 79.77          & \textbf{445.24} & 55.71 & \textbf{70.02} & \textbf{90.31} \\
GPT-4.1             & 79.25          & 611.89 & 57.84 & 67.36 & 86.24 \\
Claude 3.7 Sonnet   & 76.23          & 1289.04 & 40.34 & 47.07 & 73.31 \\
GPT-4.1 Mini        & 70.42          & 1194.77 & 48.61 & 52.87 & 72.77 \\
\bottomrule
\end{tabular}
\end{table*}

\subsection{Evaluation Metrics}
\label{ssec:eval_metrics} 

\subsubsection{Haversine Distance}
\label{sssec:haversine_distance} 
The final question in each GeoChain sequence (Question 21) requires the model to predict the geographic coordinates (latitude, longitude) of the depicted scene. To evaluate the accuracy of these specific predictions, we compute the \textit{Haversine distance}: the shortest distance over the Earth's surface between the predicted and ground-truth coordinates, assuming a spherical Earth. A detailed explanation of the Haversine distance calculation is provided in Section~\ref{sec:hd}. 

\subsubsection{Pass Score}
\label{sssec:pass_score} 
The \textbf{Pass Score} is computed as the average fraction of questions correctly answered across the full 21-step reasoning chain for each image. A prediction for any question is considered correct if it matches the ground-truth answer for that specific question, accounting for its type (e.g., exact match for free-text, class match for multiclass, or binary match). Crucially, for the final latitude and longitude prediction (Question 21), a response is deemed correct contributing to the Pass Score if its Haversine distance (as defined in Section~\ref{sssec:haversine_distance}) from the ground truth is less than 50km.

\subsection{Overall Model Performance}
Overall model performance (Table \ref{tab:model_level}) offers nuanced insights into current MLLM geographic reasoning. The leading Gemini models exhibit specialized strengths: Gemini-2.5-pro excels in complex multi-step reasoning (pass score 81.84\%), whereas Gemini-2.5-Flash achieves superior localization precision (445.24 km mean error), hinting at differing architectural or training optimizations. This divergence underscores that broad inferential ability and precise geolocalization are distinct skills, likely requiring separate optimization pathways rather than being monolithic capabilities. GPT-4.1 maintains a competitive position; however, the substantial localization inaccuracies of Claude 3.7 Sonnet (1289.04 km error) and GPT-4.1 Mini (1194.77 km error) underscore that robust geospatial grounding is a significant developmental hurdle, indicating a key area for advancement in MLLM capabilities.

The introduction of threshold-based localization accuracies - at City (< 25 km), Region (< 200 km), and Country (< 750 km) levels further refines this performance landscape. Gemini-2.5-pro's superior performance is reinforced by its top-tier City-level precision (59.38\%). Complementing this, Gemini-2.5-Flash excels in broader accuracy, leading at both Region-level (70.02\%) and Country-level (90.31\%). GPT-4.1 also demonstrates notable strength in City-level performance (57.84\%), surpassing Gemini-2.5-Flash in this specific high-precision context. Conversely, Claude 3.7 Sonnet's previously noted localization challenges are starkly emphasized by its profound difficulties at these finer scales (e.g., 40.34\% at City-level), performing below GPT-4.1 Mini (48.61\% at City-level) here. These granular metrics effectively highlight that achieving reliable, high-confidence City-level precision is a primary differentiator and a significant challenge across the evaluated MLLMs.

\begin{table}[ht]
\centering
\caption{Performance by image difficulty. Accuracy (\%) and Haversine distance (km) for each difficulty level.}
\label{tab:image_difficulty}
\setlength{\tabcolsep}{2pt}
\begin{tabular}{llcc}
\toprule
\textbf{Model} & \textbf{Diff} & \textbf{Pass Score} & \textbf{M. Dist.} \\
\midrule
Claude 3.7    & Easy    & 77.2  & 885.86  \\
   Sonnet                 & Medium  & 78.3  & 989.13  \\
                    & Hard    & 73.2  & 2000.14 \\
\midrule
GPT-4.1         & Easy    & 70.8  & 863.19  \\
Mini                 & Medium  & 73.2  & 827.78  \\
                    & Hard    & 67.3  & 1910.44 \\
\midrule
GPT-4.1             & Easy    & 79.3  & 357.36  \\
                    & Medium  & 81.6  & 428.46  \\
                    & Hard    & 76.8  & 1052.13 \\
\midrule
Gemini-2.5    & Easy    & 80.5  & \textbf{287.61}  \\
   Flash                 & Medium  & 82.5  & \textbf{188.45} \\
                    & Hard    & 76.3  & 873.78  \\
\midrule
Gemini-2.5& Easy    & \textbf{83.3}  & 300.29  \\
         Pro           & Medium  & \textbf{84.2}  & 304.32 \\
                    & Hard    & \textbf{78.0}  & \textbf{866.62}  \\
\bottomrule
\end{tabular}
\end{table}

\subsection{Breakdown by Image Difficulty}
Analyzing model performance by image difficulty (Table \ref{tab:image_difficulty}) reveals critical operational characteristics. As expected, 'Hard' images significantly challenge all models, leading to substantial increases in mean localization errors often exceeding 1000-2000 km for several models. The Gemini models consistently lead: Gemini-2.5-pro achieves top Pass Scores across all difficulties (e.g., 78.0\% on Hard), while Gemini-2.5-Flash generally provides superior localization on 'Easy' and 'Medium' images (e.g., 188.45 km on Medium). Notably, Gemini-2.5-pro performs the best for localization precision on 'Hard' images (866.62 km), possibly where its stronger inferential capacity becomes decisive. An intriguing anomaly is the better localization by some models, like Gemini-2.5-Flash, on 'Medium' versus 'Easy' images, potentially due to bias towards certain cities in pre-training data. Furthermore, Claude 3.7 Sonnet's performance is particularly interesting: despite reasonable Pass Scores (e.g., 73.2\% on Hard), its poor localization (2000.14 km on Hard) highlights a profound disconnect between understanding cues and grounding them spatially. 

\begin{figure*}[t!]
\centering
\includegraphics[width=\linewidth]{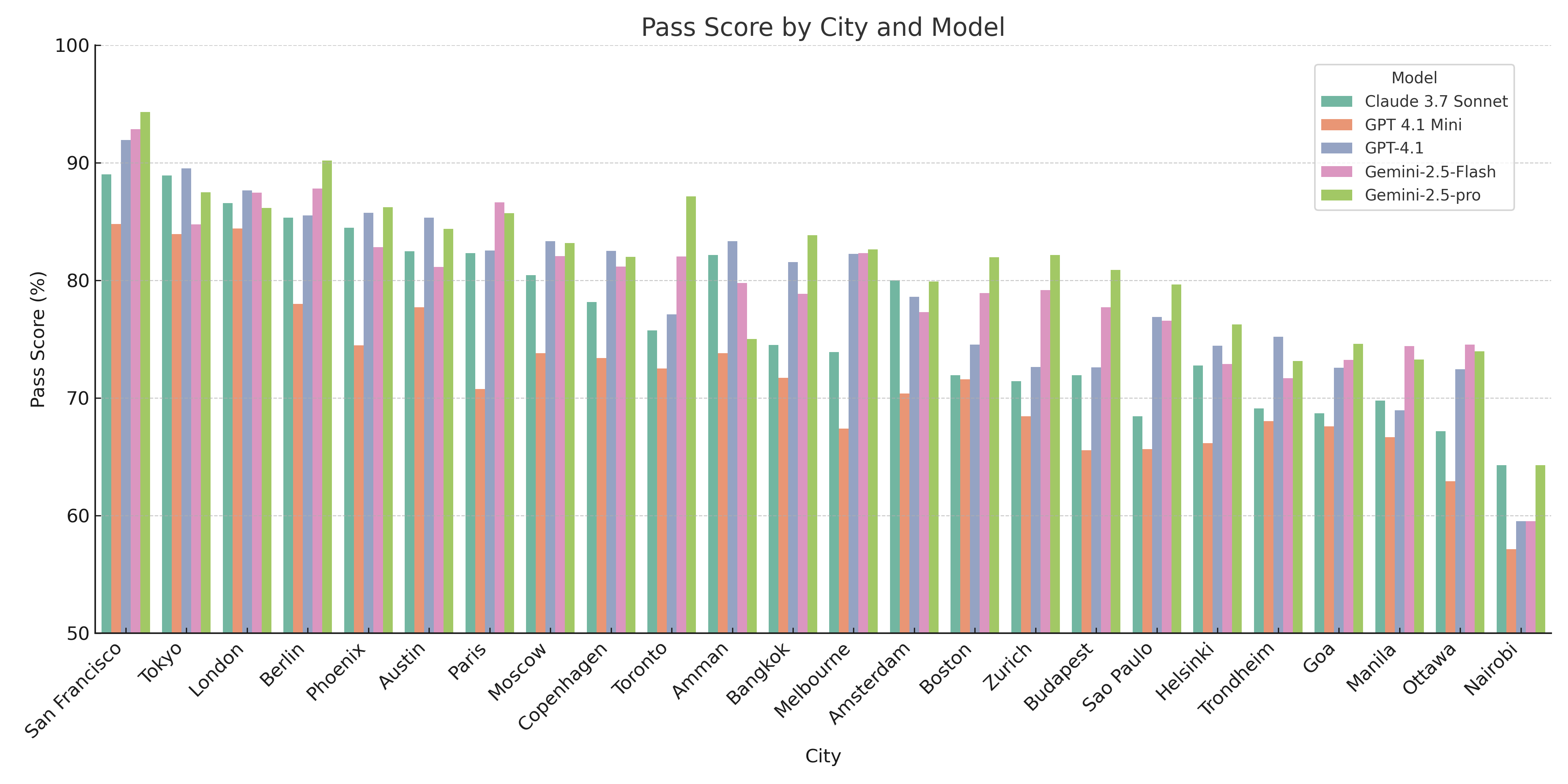}
\caption{Pass score (\%) by city, highlighting the influence of geographical location on model accuracy.}
\label{fig:city_accuracy}
\end{figure*}

\subsection{Breakdown by Question Category}
\begin{table*}[ht]
\centering
\caption{Pass score (\%) by question category.}
\label{tab:question_category}
\begin{tabular}{lccccc}
\toprule
\textbf{Model} & \textbf{Visual} & \textbf{Terrain} & \textbf{Geo Localization} & \textbf{Cultural} & \textbf{Exact Loc.} \\
\midrule
Claude 3.7 Sonnet   & \textbf{92.8} & 84.7 & 69.4 & 67.4 & 51.0 \\
GPT-4.1 Mini        & 92.3 & 78.7 & 64.1 & 56.8 & 40.7 \\
GPT-4.1             & 91.8 & 84.8 & \textbf{76.9} & 68.3 & 61.5 \\
Gemini-2.5-Flash    & 92.4 & 86.0 & 73.5 & 75.3 & 59.8 \\
Gemini-2.5-pro      & 92.1 & \textbf{87.4} & 76.8 & \textbf{77.9} & \textbf{63.5} \\
\bottomrule
\end{tabular}
\end{table*}
Analyzing Pass Scores by question category (Table \ref{tab:question_category}), informed by the benchmark's diverse question structures (e.g., visual queries versus free-text specific knowledge), reveals distinct performance strata. Foundational "Visual" questions, focusing on direct object presence (e.g., "Do you see any boats?"), yield universally high scores (all models >91.8\%), suggesting robust basic visual grounding and low immediate hallucination, with Claude 3.7 Sonnet leading (92.8\%). Similarly, "Terrain" identification is generally strong. In contrast, categories like "Geo Localization" and "Cultural" show mixed results; models likely handle simpler, coarse queries (e.g., continent identification) better than challenging free-text questions requiring specific knowledge (e.g., city/state names, language identification). Unsurprisingly, "Exact Loc" demanding precise latitude/longitude output—is definitively the most challenging category across all models. Within this landscape, Gemini-2.5-pro consistently excels, particularly in the more demanding categories like "Terrain" (87.4\%), "Cultural" (77.9\%), and "Exact Loc." (63.5\%). GPT-4.1 also demonstrates strong performance, notably in "Geo Localization" (76.9\%) and "Exact Loc." (61.5\%). Claude 3.7 Sonnet's profile, with its excellent "Visual" scores but significantly weaker "Exact Loc." performance (51.0\%), starkly illustrates a common theme: a disconnect between initial cue processing and final, precise geospatial grounding, which remains the primary MLLM hurdle.

\subsection{Breakdown by City}

A city-level view (Fig. \ref{fig:city_accuracy}) shows that performance is far from uniform:

Gemini-2.5-pro is the most stable, topping the leaderboard in 20 / 24 cities and exceeding 85\% accuracy in visually distinctive urban centres such as Tokyo, Zurich and Toronto. Gemini-2.5-Flash and GPT-4.1 follow closely, maintaining more than \textbf{75\%} accuracy in most regions. Performance on Claude 3.7 Sonnet and GPT 4.1 Mini fluctuate sharply: they perform competitively in cue-rich European cities (Paris, Berlin) but collapse in visually ambiguous locales (Nairobi, São Paulo, Amman). Mean Haversine error (Fig. \ref{fig:city_distance}) confirms the pattern: Gemini-2.5-pro keeps errors below 300 km in nearly every city, whereas Claude and GPT 4.1 Mini exceed 1000 km in several cases (Helsinki, Melbourne, São Paulo).

These results highlight how regional factors such as vegetation, signage language, traffic orientation and architectural style strongly modulate geolocation accuracy.

\section{Conclusion}
\label{sec:conclusion}
This paper introduced GeoChain, a large-scale, chain-of-thought benchmark designed to dissect multimodal geographic reasoning in MLLMs using street-level imagery and a 21-step diagnostic framework. Our evaluations on the curated GeoChain Test-Mini subset reveal that even leading MLLMs exhibit significant deficiencies in visual grounding, reasoning consistency, and localization accuracy, particularly as task and visual complexity escalate. By enabling a granular, step-by-step analysis, GeoChain moves beyond simple end-task accuracy to pinpoint these critical failure modes, thereby providing an essential diagnostic resource and methodology. We anticipate that GeoChain will steer future research towards developing more robust, geographically aware, and reliable AI systems capable of nuanced real-world understanding.

\section{Limitations}
\label{sec:limitations}
While GeoChain offers a novel diagnostic approach, we acknowledge several limitations. GeoChain is built upon the Mapillary Street-Level Sequences training split; consequently, while our chain-of-thought reasoning framework and the overall task are novel, there is a potential that MLLMs have encountered these specific visual scenes or highly similar ones during their extensive pre-training. Evaluating performance on truly "unseen" street-level imagery is an inherent challenge for the field, given the ubiquity of data from sources like Google Street View \cite{5481932} and OpenStreetMap \cite{Haklay2008OpenStreetMap}, meaning that performance assessments may partly reflect familiarity with certain visual data rather than solely generalization to entirely new scenes. Additonally, the underlying geographical distribution of the images, though diverse, retains some skew, potentially affecting the generalizability of the findings in all urban contexts. Furthermore, our locatability score's precision is contingent upon the accuracy of an upstream semantic segmentation model, which could introduce noise into the difficulty stratification.

\section{Usage of Generative AI tools}
We utilized Generative AI tools to help improve the language, phrasing, and readability of this manuscript.

\nocite{Jadhav_Cao_Shetty_Kumar_Sharma_Sukboontip_Tamarapalli_Zhang_Koul_2025}
\nocite{jain2023maeamultimodalattributionembodied}
\nocite{yerramilli2024attributionregularizationmultimodalparadigms}
\nocite{yerramilli2024semanticaugmentationimagesusing}
\nocite{9412739}
\nocite{caoembodied}
\nocite{grover2025huemanityprobingfinegrainedvisual}
\nocite{phan2025humanitysexam}
\bibliography{custom}

\appendix

\section{Appendix}
\label{sec:appendix}

\subsection{Implementation Details}
\subsubsection{Questions}
\label{sec:questions}
This section details the complete 21-question sequence (Table \ref{tab:geochain_questions_detailed_std}) that forms the core of the GeoChain benchmark, designed to evaluate the step-by-step geographic reasoning capabilities of Multimodal Large Language Models (MLLMs).  Each question in the sequence is characterized by its rank, designated difficulty level (Easy, Medium, or Hard), expected response format (Binary, Multiclass, or Free-text), and its primary Question Category (Visual Cues, Geographical localization, Culture/Infrastructure, Terrain/Environment, or Exact Location). This comprehensive listing provides a transparent foundation for understanding the specific tasks underpinning the performance evaluations discussed throughout this paper.

\subsubsection{System Prompt}
To guide the Multimodal Large Language Models (MLLMs) and standardize their responses for the GeoChain benchmark tasks, the following system prompt was consistently employed:

\begin{tcolorbox}[title=System Prompt]
You are an accurate geolocation model. Given the image, answer the following questions in order. Please provide your best guess. Each question is also provided with question type. For Binary questions, answer Yes/No only. For Multiclass questions, answer as one of the provided options in brackets. Final question type is a free text question, answer it as a free string text. If you are not sure about the answer, give your best guess. Answer format should be a json dict with question indices as keys (0 indexed) and values as Answer: <answer>, Reasoning: <reasoning>.
\end{tcolorbox}

\subsubsection{Tools and Infrastructure}
The execution of model inference was managed by Promptfoo\footnote{\url{https://www.promptfoo.dev}}, a platform that ensures reproducibility in benchmarking by offering versatile prompt configuration and effective API linkage. We used the transformers library in Hugging Face\footnote{\url{https://huggingface.co/docs/transformers/en/index}}; to run the MaskFormer model for computing segmentation masks. These calculations were performed on an NVIDIA GeForce RTX 3060 graphics processing unit.
\begin{table*}[htbp] 
\centering
\caption{The GeoChain 21-Step Benchmark Question Set.}
\label{tab:geochain_questions_detailed_std}
\footnotesize 
\setlength{\tabcolsep}{4pt} 

\begin{tabular}{@{}c l p{5.5cm} p{3cm} p{3cm}@{}} 
\toprule
\textbf{Rank} & \textbf{Difficulty} & \textbf{Question} & \textbf{Question Type} & \textbf{Question Category} \\
\midrule
1 & Easy & Do you see any boats or ships? & Binary & Visual Cues \\
2 & Easy & Do you see one or more of the following vehicles: Bus, Truck, Car, Van, Motorbike, Minibike, Bicycle? & Binary & Visual Cues \\
3 & Easy & Can you see any traffic lights? & Binary & Visual Cues \\
4 & Easy & Can you see any flag? & Binary & Visual Cues \\
5 & Easy & Would you say this location is near the Equator? & Binary & Geographical localization \\
6 & Easy & Does this location seem to be close to the Poles? & Binary & Geographical localization \\
7 & Easy & Is this place located in the Northern Hemisphere? & Binary & Geographical localization \\
8 & Easy & Which continent best describes where this location is? (7 continents: North America/South America/Europe/Africa/Asia/Oceania/Antarctica) & Multiclass  & Geographical localization \\
9 & Medium & What side of the road do vehicles drive on here? (Left/Right) & Multiclass & Culture/Infrastructure \\
10 & Medium & What country is this place located in? & Free-text & Geographical localization \\
11 & Medium & Is this place near coast? & Binary & Terrain/Environment \\
12 & Medium & Does this location appear to be an island? & Binary & Terrain/Environment \\
13 & Easy & Is this place located in a desert region? & Binary & Terrain/Environment \\
14 & Easy & Does this location seem to be in a mountainous or hilly region? & Binary & Terrain/Environment \\
15 & Medium & What is the most likely climate type for this location? (5 main climate types: Tropical/Dry/Temperate/Continental/Polar) & Multiclass  & Terrain/Environment \\
16 & Easy & Does this place look like a big city? & Binary & Culture/Infrastructure \\
17 & Medium & Would you classify this place as a small town? & Binary & Culture/Infrastructure \\
18 & Hard & What language(s) are most likely spoken at this place? & Free-text & Culture/Infrastructure \\
19 & Hard & Can you name the state or province this place belongs to? & Free-text & Geographical localization \\
20 & Hard & What is the name of the city, town, or village seen here? & Free-text & Geographical localization \\
21 & Hard & Based on everything observed, what are the latitude and longitude coordinates of this place? Please give a tuple of float coordinates (lat, lon) & Free-text & Exact Location \\
\bottomrule
\end{tabular}
\end{table*}
\subsection{Haversine Distance}
\label{sec:hd}
Haversine Distance the shortest distance over the Earth's surface between the predicted and ground-truth coordinates, assuming a spherical Earth. 

The Haversine formula is given by:
$$
\Delta \phi = \phi_2 - \phi_1 \notag \\
$$
$$
\Delta \lambda = \lambda_2 - \lambda_1 \notag \\
$$
$$
a = \sin^2\left(\frac{\Delta \phi}{2}\right) + \cos(\phi_1) \cos(\phi_2) \sin^2\left(\frac{\Delta \lambda}{2}\right)
$$
$$
d = 2R \cdot \arcsin\left(\sqrt{a}\right)
$$

Here, $d$ is the Haversine distance between two points $(\phi_1, \lambda_1)$ and $(\phi_2, \lambda_2)$. This metric provides an interpretable and robust way to measure geographic prediction error.

\subsection{Additional Analysis}
\begin{table*}[t!]
\centering
\caption{Pass score (\%) across question difficulty and image difficulty. Each row shows performance on a given question difficulty across images of increasing ambiguity.}
\label{tab:image_qdiff_matrix}
\begin{tabular}{llccc}
\toprule
\textbf{Model} & \textbf{Question Difficulty} & \textbf{Easy Images} & \textbf{Medium Images} & \textbf{Hard Images} \\
\midrule
Claude 3.7 Sonnet   & Easy   & 89.3 & 89.1 & 87.7 \\
                    & Medium & 76.0 & 75.7 & 72.0 \\
                    & Hard   & 45.8 & 52.7 & 34.8 \\
\midrule
GPT 4.1 Mini        & Easy   & 86.7 & 86.7 & 84.2 \\
                    & Medium & 66.1 & 67.3 & 62.1 \\
                    & Hard   & 37.4 & 44.6 & 27.9 \\
\midrule
GPT-4.1             & Easy   & 87.9 & 87.5 & 84.3 \\
                    & Medium & 75.5 & 76.5 & 71.8 \\
                    & Hard   & 51.8 & 60.0 & 43.3 \\
\midrule
Gemini-2.5-Flash    & Easy   & 90.7 & 91.0 & 89.4 \\
                    & Medium & 76.7 & 77.8 & 74.2 \\
                    & Hard   & 47.3 & 54.9 & 38.4 \\
\midrule
Gemini-2.5-pro      & Easy   & \textbf{91.6} & \textbf{91.3} & \textbf{89.8} \\
                    & Medium & \textbf{78.2} & \textbf{79.9} & \textbf{75.7} \\
                    & Hard   & \textbf{52.4} & \textbf{61.6} & \textbf{45.9} \\
\bottomrule
\end{tabular}
\end{table*}

\subsubsection{Image Difficulty vs Question Difficulty Interaction}
To analyze how visual and reasoning difficulty interact, we compute a two-dimensional pass rate matrix over \textbf{question difficulty} (Easy, Medium, Hard) and \textbf{image difficulty} (Easy, Medium, Hard). Table~\ref{tab:image_qdiff_matrix} presents this breakdown for each model.

\noindent
We observe a consistent trend across all models: accuracy declines with both increasing \textit{image} difficulty and \textit{question} difficulty. Importantly, hard questions on hard images represent the most challenging setting, with pass rates often below 40\%—even for state-of-the-art models.

Gemini-2.5-pro shows the strongest resilience across the board, maintaining high scores even on hard questions in ambiguous scenes. In contrast, Claude 3.7 Sonnet and GPT 4.1 Mini exhibit large drops in performance under compounding difficulty, confirming their brittleness in multi-factor reasoning.

This matrix allows us to quantify model \textit{sensitivity to visual ambiguity} and pinpoint failure modes. For example, a model that performs well on hard questions from easy images but poorly on the same questions from hard images may lack robustness in interpreting noisy visual cues. Conversely, a model that fails uniformly on hard questions indicates weaknesses in logical chaining or symbolic inference.
Together, this analysis emphasizes the need for benchmarks that probe cross-modal interactions, rather than evaluating visual or linguistic difficulty in isolation.

\subsubsection{Breakdown by Question Difficulty}

To better understand how models handle increasing reasoning complexity, we group questions by their annotated difficulty levels: \textbf{Easy}, \textbf{Medium}, and \textbf{Hard}. These difficulty tags were assigned manually based on the subtlety, required external knowledge, and ambiguity of each question.

\begin{table}[ht]
\centering
\caption{Pass score (\%) by question difficulty.}
\label{tab:question_difficulty}
\begin{tabular}{lccc}
\toprule
\textbf{Model} & \textbf{Easy} & \textbf{Medium} & \textbf{Hard} \\
\midrule
Claude 3.7 Sonnet   & 88.7 & 74.6 & 44.5 \\
GPT 4.1 Mini        & 85.9 & 65.2 & 33.4 \\
GPT-4.1             & 87.3 & 75.8 & 54.7 \\
Gemini-2.5-Flash    & 90.8 & 76.2 & 51.3 \\
Gemini-2.5-pro      & \textbf{91.1} & \textbf{78.4} & \textbf{55.1} \\
\bottomrule
\end{tabular}
\end{table}
Across all models, accuracy decreases consistently with question difficulty. Gemini-2.5-pro achieves the highest pass rates at all levels, followed closely by Gemini-2.5-Flash and GPT-4.1. Interestingly, Claude 3.7 Sonnet and GPT 4.1 Mini both exhibit sharp drops on hard questions, with performance falling below 45\% and 35\%, respectively.

These findings suggest that while many models can answer surface-level geographic questions accurately, their reasoning falters as complexity increases especially when fine-grained localization or symbolic inference is required. The relatively better performance of Gemini-2.5-pro on hard questions indicates more stable multi-hop reasoning or greater robustness to subtle visual signals.

\subsubsection{Accuracy vs. Reasoning Depth}
\begin{figure}[htpb]
\centering
\includegraphics[width=\linewidth]{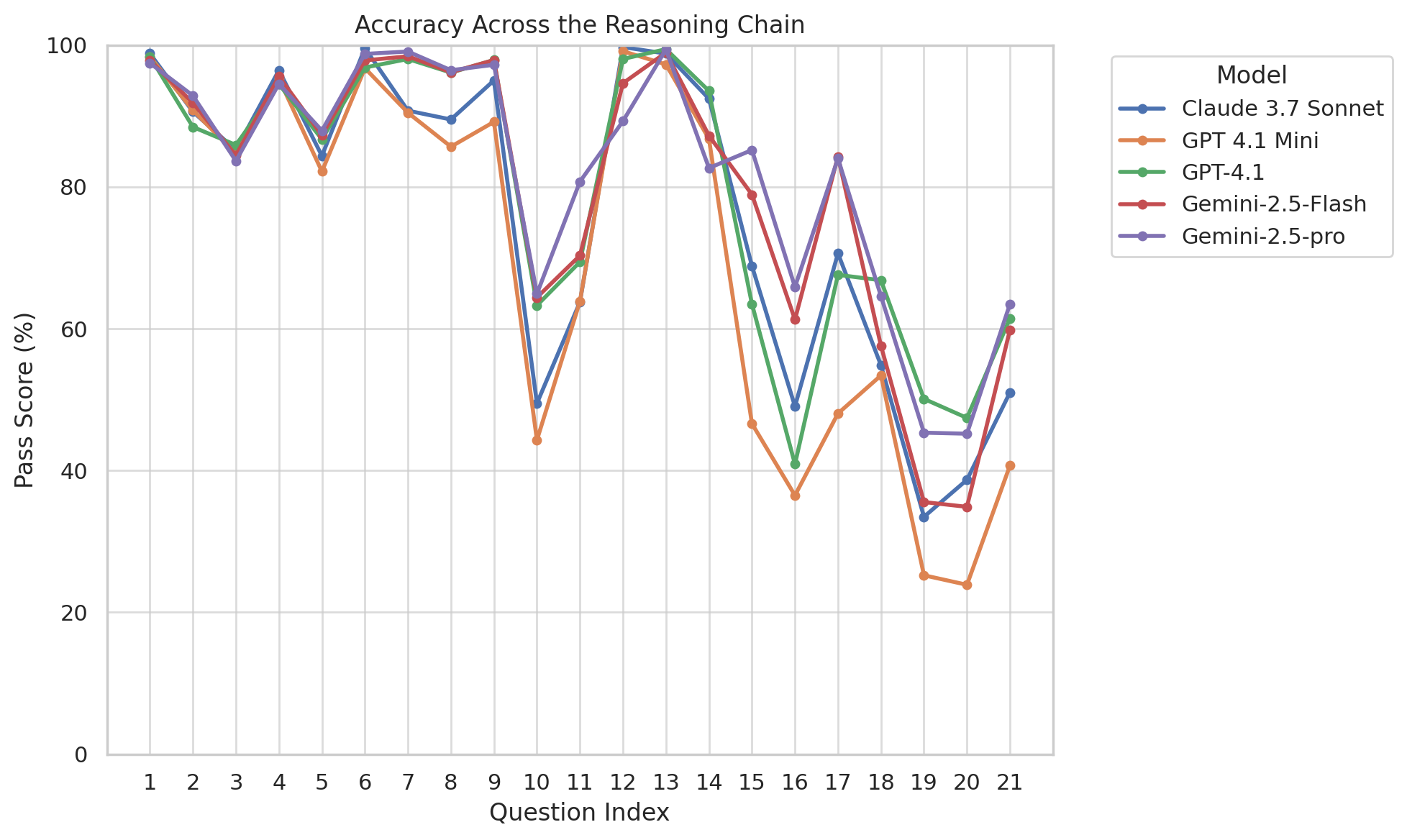}
\caption{Average pass score across the 21-step Geochain reasoning chain. Accuracy decreases as questions progress from coarse global inference to fine-grained localization.}
\label{fig:qidx_accuracy}
\end{figure}

Figure~\ref{fig:qidx_accuracy} reveals a typical degradation pattern: All models perform well in the initial questions (1–9), which ask about visual or global cues such as vehicles, hemisphere, or continent. These are relatively easy to infer on the basis of surface-level features.

As the questions become more complex and semantically demanding, the accuracy drops sharply, especially at questions 10 and 17. These questions requiring nuanced interpretation of environmental and infrastructure signals.

In particular, we observe a performance bump around questions 12–14. Despite appearing later in the sequence, these questions ask about relatively easy visual features (e.g., desert, hills, or city size). This reinforces the value of structuring questions not just by logical sequence but also by measured difficulty, allowing finer-grained diagnostics of model capability.

The final steps of the chain (questions 18–21) see the steepest drop in performance, as models are asked to predict language, administrative region, city name, and exact coordinates - tasks that require multi-modal reasoning, robust world knowledge, and low-level visual grounding.

This progressive breakdown highlights GeoChain’s utility as a diagnostic benchmark. By tracking model accuracy at each reasoning step, researchers can isolate failure modes (e.g. visual hallucination vs. failure to capture cultural cues) and understand how performance degrades under deeper spatial inference chains.

\subsubsection{Breakdown by Question Type}

\begin{figure*}[t!]
\centering
\includegraphics[width=\linewidth]{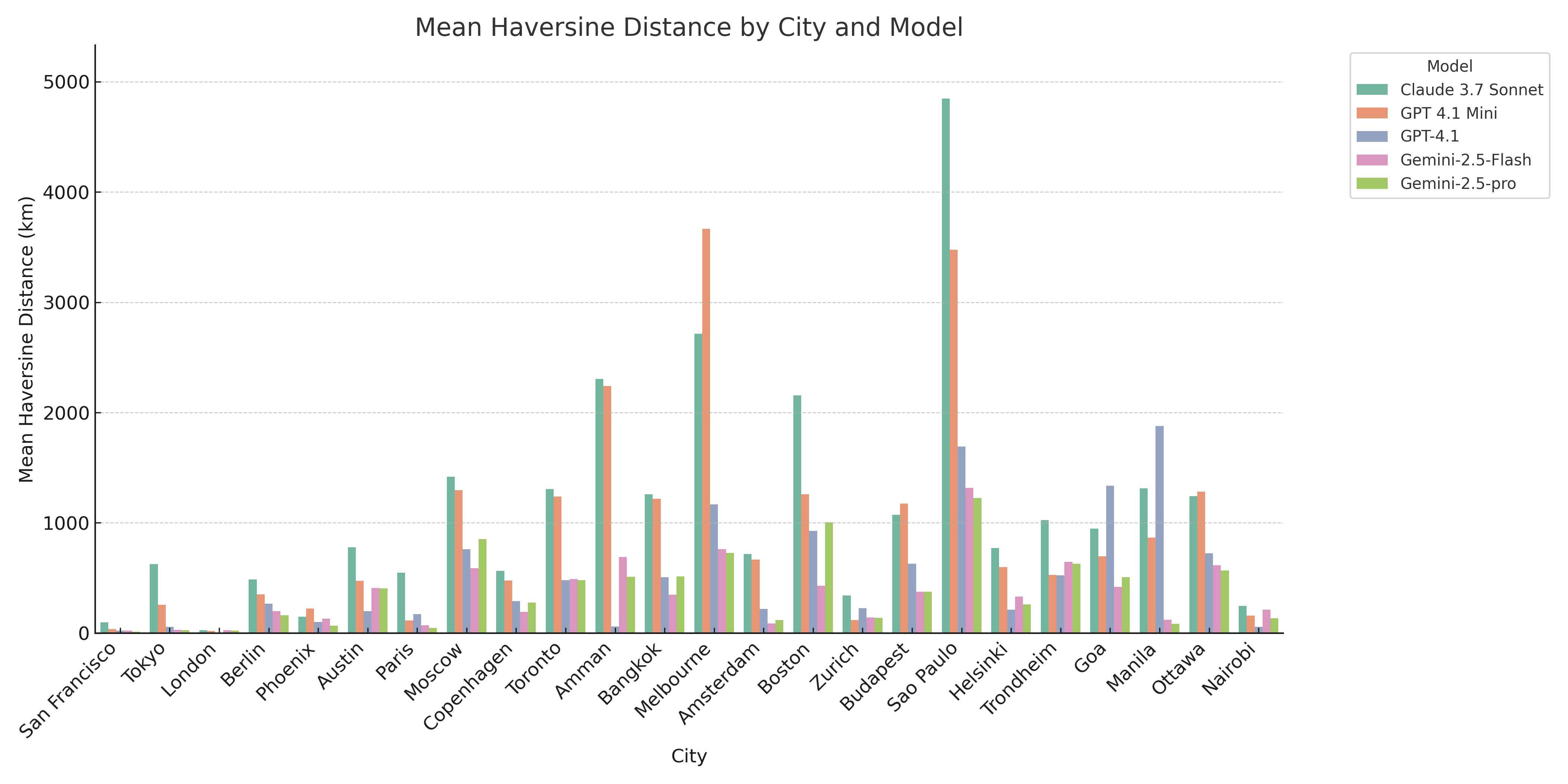}
\caption{Mean Haversine distance (km) by city and model. Larger values indicate poor localization precision.}
\label{fig:city_distance}
\end{figure*}

\begin{figure}[t]
    \centering
    \includegraphics[width=1.0\linewidth]{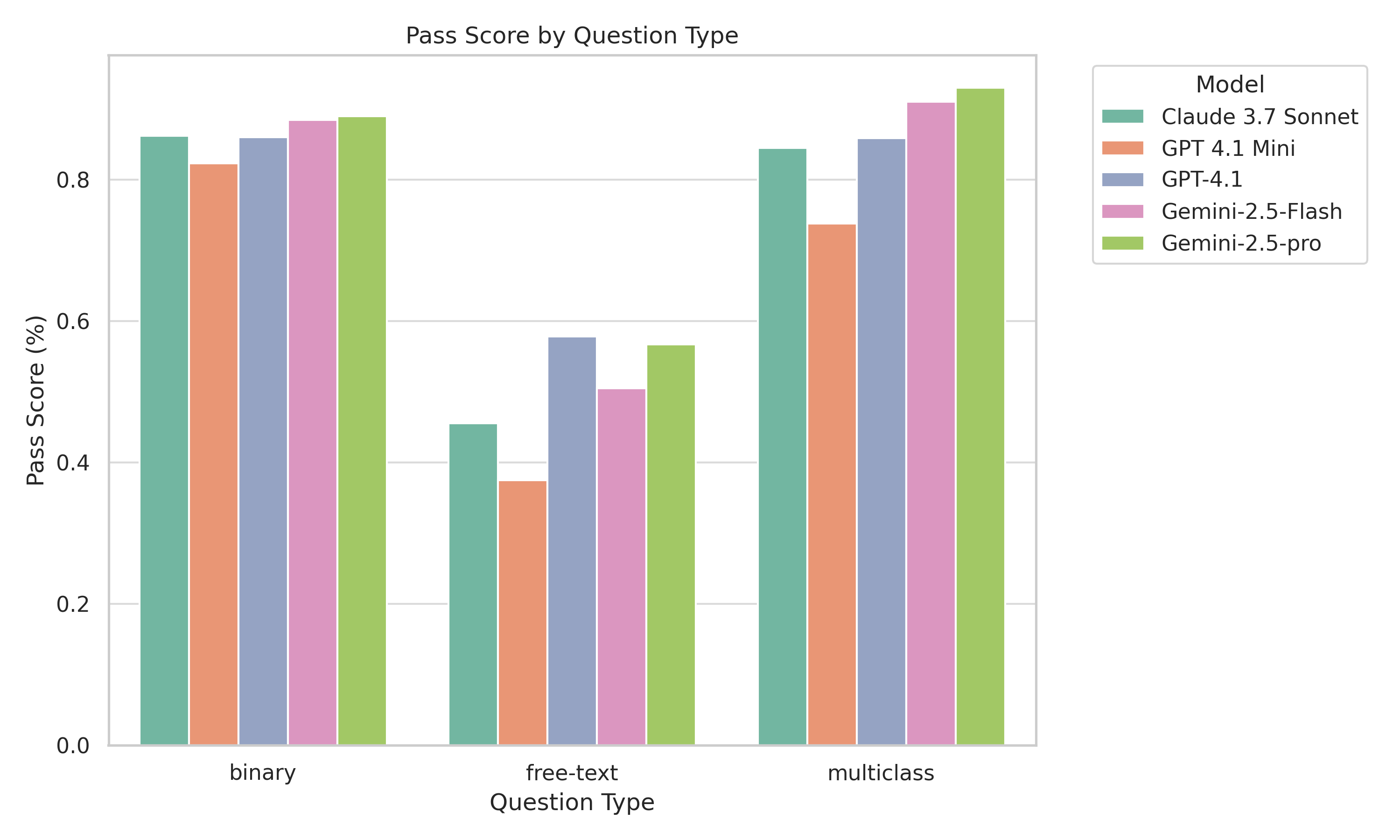}
    \caption{Model vs Question Type}
    \label{fig:model_ques_type}
\end{figure}

To assess how models handle varying degrees of response constraint, we analyzed Pass Scores across three fundamental question types: Binary, Multiclass, and Free-text, with results presented in Table \ref{tab:question_type} and Figure \ref{fig:model_ques_type}. This breakdown reveals a distinct performance hierarchy directly correlated with the open-endedness of the required answer.

Across all evaluated MLLMs, a clear difficulty gradient was observed: Binary questions yielded the highest success rates, followed by Multiclass questions, with Free-text questions proving to be the most challenging by a substantial margin. For instance, Gemini-2.5-pro achieved 88.9\% on Binary and an exceptional 92.9\% on Multiclass questions, but its score dropped to 56.7\% for Free-text tasks. This pattern of significantly lower performance on Free-text questions was universal, underscoring the inherent difficulty in precise, open-ended generation and factual recall compared to selecting from constrained options.

In the structured formats, Gemini-2.5-pro consistently led, achieving the top scores for both Binary (88.9\%) and Multiclass (92.9\%) questions, with Gemini-2.5-Flash also performing strongly. Notably, for the more demanding Free-text questions, GPT-4.1 emerged as the top performer with a Pass Score of 57.8\%, slightly ahead of Gemini-2.5-pro (56.7\%). This suggests a particular strength in GPT-4.1's generative capabilities for unconstrained answers. Claude 3.7 Sonnet demonstrated robust performance on Binary (86.2\%) and Multiclass (84.5\%) questions, often comparable to GPT-4.1, but its accuracy significantly declined on Free-text questions (45.5\%), reaffirming its challenges with precise, unprompted generation. As anticipated, GPT-4.1 Mini generally recorded the lowest scores across all types. This analysis by question type effectively highlights that while current MLLMs are largely proficient with constrained-choice tasks, open-ended free-text responses remain a key area for improvement.

\begin{table}[ht]
\centering
\caption{Pass score (\%) by question type.}
\label{tab:question_type}
\setlength{\tabcolsep}{3pt} 
\begin{tabular}{lccc}
\toprule
\textbf{Model} & \textbf{Binary} & \textbf{Multiclass} & \textbf{Free-text} \\
\midrule
Claude 3.7 Sonnet    & 86.2 & 84.5 & 45.5 \\
GPT-4.1 Mini         & 82.3 & 73.8 & 37.5 \\
GPT-4.1              & 85.9 & 85.8 & 57.8 \\
Gemini-2.5-Flash     & 88.5 & 90.9 & 50.5 \\
Gemini-2.5-pro       & \textbf{88.9} & \textbf{92.9} & \textbf{56.7} \\
\bottomrule
\end{tabular}
\end{table}

\end{document}